\documentclass[accepted]{article}
\usepackage{amsmath,amssymb}

\usepackage{microtype}
\usepackage{graphicx}
\usepackage{subfigure}
\usepackage{booktabs} % for professional tables
\usepackage{appendix}
\usepackage{numprint}
\npthousandsep{,}

\usepackage[]{icml2021}

\icmltitlerunning{Mode Exploration in Bayesian Model Averaging for Neural Networks}

\begin{document}

\twocolumn[
\icmltitle{On the Effectiveness of Mode Exploration in Bayesian Model Averaging for Neural Networks}



\icmlsetsymbol{equal}{*}

\begin{icmlauthorlist}
\icmlauthor{John T. Holodnak}{to}
\icmlauthor{Allan B. Wollaber}{to}
% \icmlauthor{Bauiu C.~Yyyy}{equal,to,goo}
% \icmlauthor{Cieua Vvvvv}{goo}
% \icmlauthor{Iaesut Saoeu}{ed}
% \icmlauthor{Fiuea Rrrr}{to}
% \icmlauthor{Tateu H.~Yasehe}{ed,to,goo}
% \icmlauthor{Aaoeu Iasoh}{goo}
% \icmlauthor{Buiui Eueu}{ed}
% \icmlauthor{Aeuia Zzzz}{ed}
% \icmlauthor{Bieea C.~Yyyy}{to,goo}
% \icmlauthor{Teoau Xxxx}{ed}
% \icmlauthor{Eee Pppp}{ed}
\end{icmlauthorlist}

\icmlaffiliation{to}{Massachusetts Institute of Technology, Lincoln Laboratory}
% \icmlaffiliation{goo}{Googol ShallowMind, New London, Michigan, USA}
% \icmlaffiliation{ed}{School of Computation, University of Edenborrow, Edenborrow, United Kingdom}

\icmlcorrespondingauthor{John Holodnak}{john.holodnak@ll.mit.edu}

\icmlkeywords{Machine Learning, ICML}

\vskip 0.3in
]



\printAffiliationsAndNotice{}  % leave blank if no need to mention equal contribution

\begin{abstract}
Multiple techniques for producing calibrated predictive probabilities using deep neural networks in supervised learning settings have emerged that leverage approaches to ensemble diverse solutions discovered during cyclic training or training from multiple random starting points (deep ensembles).  However, only a limited amount of work has investigated the utility of exploring the local region around each diverse solution (posterior mode).  Using three well-known deep architectures on the CIFAR-10 dataset, we evaluate several simple methods for exploring local regions of the weight space with respect to Brier score, accuracy, and expected calibration error. We consider both Bayesian inference techniques (variational inference and Hamiltonian Monte Carlo applied to the softmax output layer) as well as utilizing the stochastic gradient descent trajectory near optima. While adding separate modes to the ensemble uniformly improves performance, we show that the simple mode exploration methods considered here produce little to no improvement over ensembles without mode exploration.

%This document provides a basic paper template and submission guidelines.
%Abstracts must be a single paragraph, ideally between 4--6 sentences long.
%Gross violations will trigger corrections at the camera-ready phase.
\end{abstract}

\section{Introduction}

Neural networks have been successfully used in many application areas, but perhaps most prominently in image classification.  Unfortunately, as networks have become deeper and more complex, their probabilistic outputs have become less calibrated \cite{Guo2017}, and as a result, they do not provide a useful quantification of predictive uncertainty.  Many techniques have been proposed to alleviate this problem, including Bayesian Neural Networks \cite{MacKay1992, Graves2011}, deep ensembles \cite{Lakshminarayanan2017}, Monte Carlo dropout \cite{Gal2016}, and temperature calibration \cite{Guo2017}.  

\citet{Wilson2020}, attempt to unify understanding of techniques that aggregate predictions from neural networks with different weights through the lens of the Bayesian Model Average (BMA) \cite{Hoeting1999}
$$
p(y|x;D) = \int p(y|x;\theta)p(\theta|D) d\theta.
$$
In the above, $y \in Y=\{1, \hdots, C\}$ is the true label for data point $x \in X$.  $D \subset X \times Y$ is the training dataset.  For us, $p(y|x;\theta)$ represents the neural network's predicted probabilities for data point $x$, given specific network weights $\theta$, and $p(\theta|D)$ is the posterior distribution of weights, given the training dataset.  Usually, the integral over $\theta$ is ``approximated'' using a single setting of weights, typically obtained from an optimization procedure such as Stochastic Gradient Descent (SGD).  As neural networks typically have multi-modal posteriors \cite{Muller1998}, this approximation is decidedly sub-optimal.  Many researchers have proposed methods to approximate the BMA.  We now discuss a few of the prominent techniques.

To approximate the BMA, one option is to estimate the integral above using samples from the posterior distribution obtained via Markov Chain Monte Carlo (MCMC) or from a variational approximation to the posterior (often a product of independent Gaussians).  Unfortunately, MCMC scales poorly to large datasets and also has difficulty sampling from complicated posteriors in high dimensions.  Variational Inference (VI) is more tractable and is implemented in common packages like Tensorflow and PyTorch, but in its typical implementation infers a uni-modal posterior approximation.  An alternative is to apply Stochastic Gradient Langevin Dynamics (SGLD), which adds Gaussian noise according to a decaying schedule to SGD and allows sampling from the posterior distribution of parameters \cite{Welling2011}.  There have been a few recent attempts (see \citet{Wenzel2020, Izmailov2021}) to better understand the posterior distribution of Bayesian Neural Networks and whether sampling from the posterior results in improved models.

Other not-explicitly-Bayesian techniques can also be viewed as non-trivial approximations to the BMA in that they average predictions over multiple sets of weights, though these weights are obtained by methods other than sampling from the posterior or approximate posterior.  Such techniques include training multiple networks from different starting locations (deep ensembles) \cite{Lakshminarayanan2017}, saving weights from the SGD trajectory with either a traditional or cyclic learning rate schedule (snapshot ensembles) \cite{Huang2017}, training a multiple-input-multiple-output model \cite{Havasi2021} that at test time produces multiple predictions for an input, or building a Gaussian posterior approximation from the SGD trajectory \cite{Maddox2019}.

Intuitively, ensembles are effective when the individual models make diverse but accurate predictions.  To this end, ensembles of models corresponding to different posterior modes are likely to be more diverse than ensembles of models all drawn from the same mode.  This intuition is supported by the success of deep ensembles \cite{Lakshminarayanan2017}.  A recent comparison by \citet{Ovadia2019} shows that deep ensembles are more robust in terms of accuracy and calibration than solutions from explicitly Bayesian techniques such as Stochastic Variational Inference.  An interpretation of this result is that it is better to draw one ``sample'' from each of $K$ modes than to draw $K$ samples from a single mode \cite{Wilson2020}.

A logical next question is whether it is advantageous to average model predictions \textit{within} posterior modes as well as \textit{between} posterior modes.  This question is partially explored by \citet{Wilson2020}, \citet{Zhang2020}, and \citet{Tran2020}.  \citet{Wilson2020} construct Gaussian posterior approximations using the SGD trajectory initialized at several stochastic weight averaged neural network solutions\footnote{Stochastic weight averaging refers to averaging weights from multiple neural networks to arrive at a single network.}.  They then average predictions over a few weights sampled from each Gaussian posterior approximation.  They find that this approach outperforms deep ensembles as well as ensembles of stochastic weight averaged networks on corrupted CIFAR-10 images.  In addition, they show near monotonic improvement in terms of the negative log-likelihood of PreResNet20 as the number of models is increased from one to ten (for each of the methods mentioned above).  \citet{Zhang2020} utilize a cyclic learning rate to identify several posterior modes and either Stochastic Gradient Langevin Dynamics (SGLD) or Stochastic Gradient Hamiltonian Monte Carlo (SGHMC) to sample from them.  They demonstrate that SGLD and SGHMC, run in a cyclic fashion, are more effective than traditional SGLD or SGHMC, presumably because the cyclic variants sample from multiple modes.  \citet{Tran2020} parameterize each weight matrix with a rank one factor and perform variational inference for this relatively small number of weights.  They model the posterior distribution as a mixture distribution with a few components, which enables identification of multiple modes and then draw a few weight samples at test time from each mode.  They demonstrate results that are comparable to approaches using several times as many parameters on multiple datasets.

The experiments described above demonstrate that the SGD trajectory can be used to explore posterior modes and obtain improved predictive performance either by building a local Gaussian approximation or via adding noise to the gradient and thus the trajectory (SGLD and SGHMC).  In this work, we look to better understand whether the SGD trajectory can be used directly to ``sample'' from a mode and also to examine whether explicitly Bayesian techniques (VI and MCMC) operating on the last layer of the neural network (which is computationally tractable) can capture within-mode model diversity.

To be specific, we perform experiments on CIFAR-10 using two ensemble types (deep and cyclic) and several inference types: SGD; saving samples from the SGD trajectory; variational inference (VI) on the softmax output layer; and MCMC on the softmax output layer.  VI on the last layer has been studied previously, for example see \cite{Ovadia2019}.  While it was noted to not perform much differently from standard SGD in the context of a \textit{single} model, we include it due to the simplicity of implementing it in Tensorflow and to provide a baseline for the performance of MCMC on the last layer.  Because MCMC scales poorly with dataset size, we accelerate it using Bayesian coresets \cite{Huggins2016, Campbell2019}.  We provide more detail on this acceleration in Appendix \ref{sec:BC}.  To our knowledge, this is the first use of coresets in the context of accelerating MCMC for neural networks.  We are motivated to explore MCMC in the context of deep neural networks by the discussion section in \cite{Zhang2020} and both \cite{Wenzel2020} and \cite{Izmailov2021}.

Our contributions are as follows:
\begin{itemize}
\item We perform a detailed set of experiments using multiple runs of the same approximation methods across  different network architectures of increasing complexity using the same dataset and learning rate schedule.
\item We show that a few epochs of training at a small constant learning rate after using the decaying learning rate schedule from \cite{Huang2017, Zhang2020} is effective at refining the SGD solution, but the trajectory is not useful for defining a diverse set of models within the mode.
\item We provide the first, to our knowledge, experimental analysis of the effectiveness of MCMC (applied to the output layer), accelerated by a data reduction technique called Bayesian coresets, at estimating the BMA.
\item We demonstrate that Bayesian inference on the last layer typically provides little to no improvement to the Brier score over SGD.  Last layer VI and last layer MCMC produce similar results overall.
\end{itemize}

\section{Approximation Methods}

In this section, we describe the methods we use to approximate the Bayesian Model Average.  For all approximation methods, except as noted below, we use a learning rate that decays according to 
$$
r_b = \frac{r_0}{2}\cos\left(\frac{\pi * b}{B}\right) + \frac{r_0}{2},
$$
where $b$ is the batch counter, $B$ is the total number of batches across all epochs, and $r_0$ is the initial learning rate.  This allows the model to initially take large steps and then converge towards a mode.  This learning rate is also used in \cite{Huang2017, Zhang2020}.

The methods considered are as follows:
\begin{itemize}
\item \textit{Deep ensembles}: We train $M$ models from different random starting points.   We approximate the BMA with 
$$
p(y|x;D) \approx \frac{1}{M}\sum_{j=1}^{M}{p(y|x;\theta_j)},
$$
where $\theta_j$ are the parameters obtained by SGD for model $j, \, 1 \leq j \leq M$.
\item \textit{Cyclic ensembles}: We run SGD for $M$ ``cycles,'' meaning that the starting point of optimization for model $j$ is the final set of parameters for model $j-1$.  This is identical to snaphsot ensembles \cite{Huang2017}, apart from the fact that we keep the solution from the end of each cycle, rather than only the last several.  We approximate the BMA as for deep ensembles.  

\item \textit{Deep (cyclic) ensembles - SGD trajectory}: We also consider applying SGD, first with the decaying learning rate and then with a small constant learning rate for the last $K$ epochs, and saving samples from the trajectory (when in the constant learning rate stage).  We denote the weights obtained from the last $K$ epochs of training each model $j$ as $\theta_{jk}, \, 1 \leq k \leq K$.  We consider two variants in which we save a single set of weights from the final epoch (SGD-t1) or the weights from the end of each of the last $K$ epochs (SGD-tK).  The BMA for SGD-t1 is then
$$
p(y|x;D) \approx \frac{1}{M}\sum_{j=1}^{M}{p(y|x;\theta_{jK})},
$$
and the BMA for SGD-tK is
$$
p(y|x;D) \approx \frac{1}{MK}\sum_{j=1}^{M}{\sum_{k=1}^K{p(y|x;\theta_{jk})}}.
$$
We view SGD-t1 as a means to obtain a more refined single solution and SGD-tK as a way to explore the mode of the likelihood.  SGD-tK can be viewed as a snapshot version of the Gaussian approximation method from \citep{Maddox2019}, except that instead of constructing a Gaussian using the SGD iterates, we use the raw iterates themselves.  \citet{Maddox2019} note the similarity to SGLD as well.

\item \textit{Deep (cyclic) ensembles - Last layer VI}: We train $M$ models from  random starting points or the end of each cycle of a cyclic learning rate schedule, performing Bayesian inference via stochastic variational inference for the softmax output layer only.  The BMA is then approximated as
$$
p(y|x;D) \approx \frac{1}{MK}\sum_{j=1}^{M}\sum_{k=1}^K{{p(y|x;
\theta_j^{*},
(\theta_j^{'})_k}}),
$$
where $\theta_j^{*}$ are the deterministic weights obtained by SGD for model $j$ and $(\theta_j^{'})_k$ are sampled from the variational approximation to the posterior distribution of the weights in the last layer of model $j$. %$q(\theta_j^{'}) \approx  p(\theta_j^{'} | D)$.

\item \textit{Deep (cyclic) ensembles - Last layer MCMC}: As in last layer VI, except that we use the variational approximations as a starting point to identify a so-called Bayesian coreset, and then run MCMC on the coreset for the parameters associated with the softmax output layer only.  See Appendix \ref{sec:BC}.  After warmup, we extract $K$ samples from the Markov chains.  The BMA is approximated as for last layer VI, except that $(\theta_j^{'})_k$ are sampled from the posterior distribution of the weights in the last layer of model $j$ (assuming the Markov Chain has converged).  %$q(\theta_j^{'}) \approx  p(\theta_j^{'} | D)$
\end{itemize}

\section{Experiments}

In this section, we describe experiments on CIFAR-10 to evaluate the methods described in the previous section.

\begin{figure*}
\centering
\includegraphics[width=6.5in]{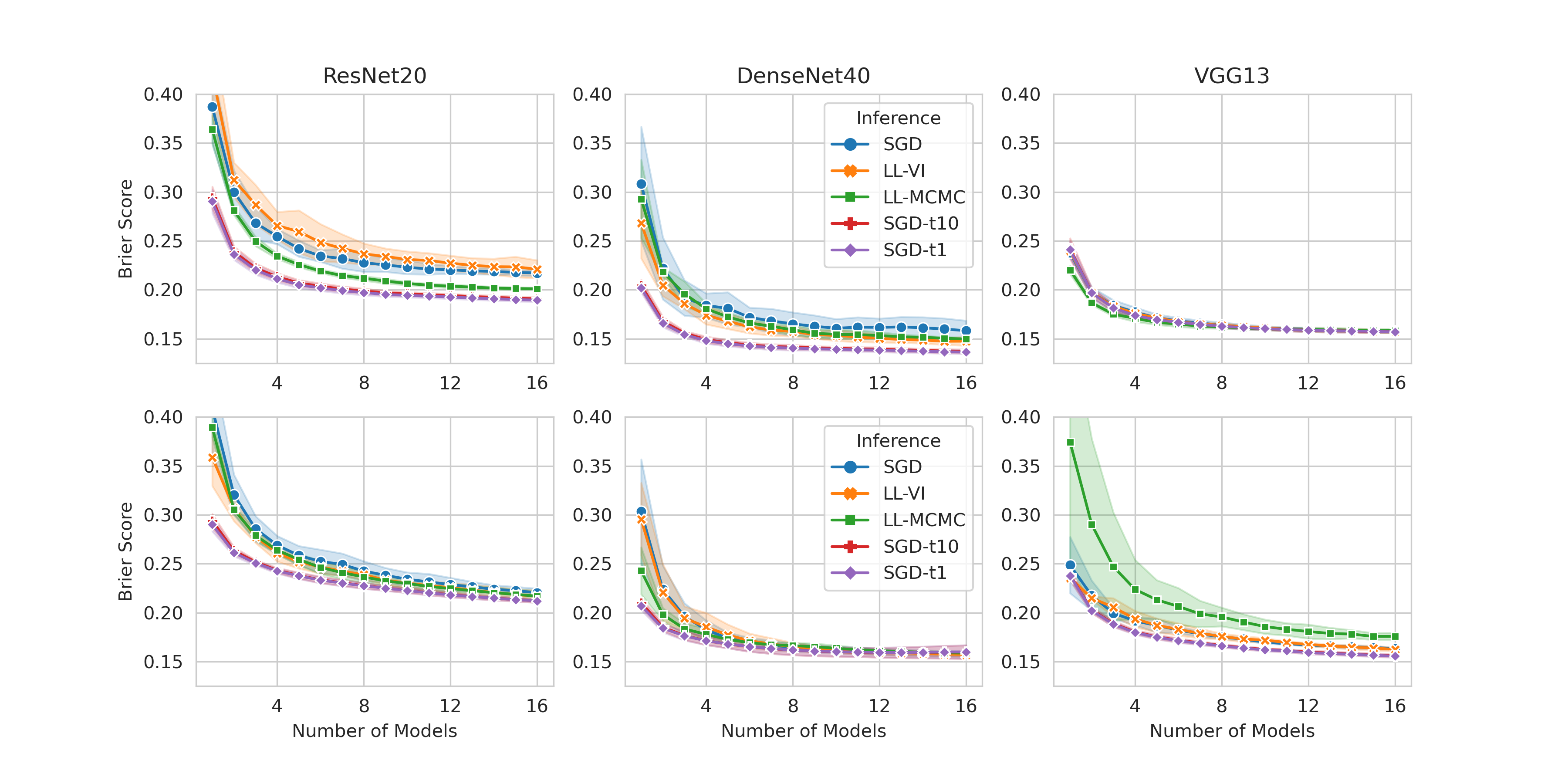}
\caption{Brier scores against number of models for different approximations to the BMA.  The top row shows deep ensembles, while the bottom row shows cyclic ensembles.  The columns show three different network architectures.  Each point represents the mean of five runs of the approximation method.  The colored region covers one standard deviation around the mean.}
\label{fig:brier}
\end{figure*}

We train networks based on VGG13, ResNet20, and DenseNet40 on CIFAR-10.  On each of the convolutional bases, we add two dense layers of size 128 and 10, with ReLU and softmax activations, respectively. We report the performance of the BMA approximation methods discussed above.  More details on training and our implementations are available in Appendix \ref{sec:CIFAR}.  In Figure \ref{fig:brier}, we display the Brier score of the ensemble models on the test set as the number of models in the ensemble increases from one to sixteen.  The Brier score is a proper scoring rule for probabilistic forecasts.  To be specific, the Brier score is defined as
$
BS = \frac{1}{N}\sum_{i=1}^N{\sum_{c=1}^C{(y_{ic} - p_{ic})^2}},
$
where $y_{ic}$ is one if item $i$ is a member of class $c$ and zero otherwise and $p_{ic}$ is the predicted probability for class $c$ on item $i$.
Note that across convolutional bases, ensemble types (deep or cyclic), and approximation methods, increasing the size of the ensemble decreases the Brier score.  

Overall, last layer VI and last layer MCMC produce results not much different from SGD, the main exceptions being ResNet20 with deep ensembles, where last layer MCMC outperforms last layer VI and SGD, and VGG13 with cyclic ensembles, where last layer MCMC performs worse than last layer VI and SGD. On the other hand, SGD-t1 and SGD-t10 perform at least as well as the other methods and considerably better on Resnet20 and DenseNet40 with deep ensembles.  Interestingly, SGD-t1 and SGD-t10 produce almost identical results, perhaps indicating that SGD by itself is ineffective at exploring modes in the likelihood.  In Figures \ref{fig:accuracy} and \ref{fig:ECE} in the Appendix, we provide similar plots, except for accuracy and expected calibration error.  Accuracy increases with the number of models almost monotonically for all approximation methods, while expected calibration error, in some cases, reaches a minimum for a small number of models (around three) and then increases. Deep ensembles overall have better (lower) Brier scores in Figure \ref{fig:brier} (top row) than the cyclic ensembles (bottom row), which is primarily due to their higher accuracies (Figure \ref{fig:accuracy}).

We view these results not as evidence that exploring modes is not useful, but as evidence that the simple techniques investigated here are not sufficient to produce within-mode model diversity.  

\section{Conclusion}

In this paper, we performed a detailed set of experiments using  several approximation methods for the BMA on three different network architectures.

One of our main goals was to answer whether simple mode exploration techniques could be leveraged to improve the BMA.  The answer to this largely seems to be negative.  Bayesian inference in the last layer (using either VI or MCMC) is not effective at improving Brier score over the baseline provided by SGD.  In addition, while it is helpful to refine the SGD solution obtained from the decreasing learning rate schedule with a few epochs of training with a constant learning rate, it does not appear helpful to save multiple solutions from those epochs.  It is possible that using a larger learning rate could make this more effective. 

We believe more work is still needed to identify the best ways to explore modes for Bayesian model averaging.  
An interesting direction would be to combine mode identification with the subspace inference technique from \cite{Izmailov2020} as a way to identify subspaces in which one could feasibly run MCMC.  Combined with the Bayesian coresets idea discussed in Appendix \ref{sec:BC}, this approach could be scalable to large datasets.

\bibliographystyle{icml2021}
\bibliography{bibliography}

\clearpage

\appendix
\section{Bayesian Coresets for Neural Networks}
\label{sec:BC}

Bayesian coresets \cite{Huggins2016, Campbell2019} is a technique to identify sparse weights for the training dataset so that the weighted log-likelihood is close to the true log-likelihood.  That is, find weights $w_n \geq 0$ such that 
$$
\sum_{i=1}^N{\ln(p(y_n|\theta))} \approx \sum_{i=1}^N{w_n\ln(p(y_n|\theta))},
$$
with the constraint that at most $N* << N$ weights are non-zero.  \citet{Campbell2018} and \citet{Campbell2019} discuss optimization techniques to solve the problem, which require computing weighted inner products of log-likelihood functions 
$$
\left<\ln(p(y_{n_i}|\theta)), \ln(p(y_{n_j}|\theta))\right>_{\pi},
$$
where $\pi$ is an approximation to the posterior distribution.
To make this computation practical, the authors utilize random projections to project the log-likelihood functions into a finite dimensional space and obtain a quick approximation to the posterior distribution to use as the weighting function (via Laplace approximation, variational inference, informed prior, etc.).

Unfortunately, MCMC also scales poorly with the number of parameters.  As a result, we only run MCMC on the last layer of the neural network.

Recall that in our work, we are attempting to explore multiple posterior modes using either deep or cyclic ensembles.  Our goal is to use MCMC to ``explore'' each mode.  As such, we need to find a coreset for each posterior mode.  We use variational inference to approximate the posterior distribution of the weights in the last layer, for each mode.  We project the likelihood functions into a \numprint{5000} dimensional space and solve the optimization problem using the technique from \cite{Campbell2018}, using a maximum coreset size of \numprint{1500}.  Finally, we run the No U-Turn Sampler (NUTS) as implemented in Tensorflow Probability with $M$ chains (one chain initialized in each mode).  We run all $M$ chains simultaneously on the same GPU.  We use \numprint{10000} warm-up iterations and then draw \numprint{1000} samples.  We thin the samples by a factor of 100 to obtain 10 samples per mode.  These samples are used in combination with the deterministic weights from the rest of the network to obtain 10 predictions per mode.

\section{CIFAR-10 details}
\label{sec:CIFAR}

\begin{figure}
\centering
\includegraphics[width=3.1in]{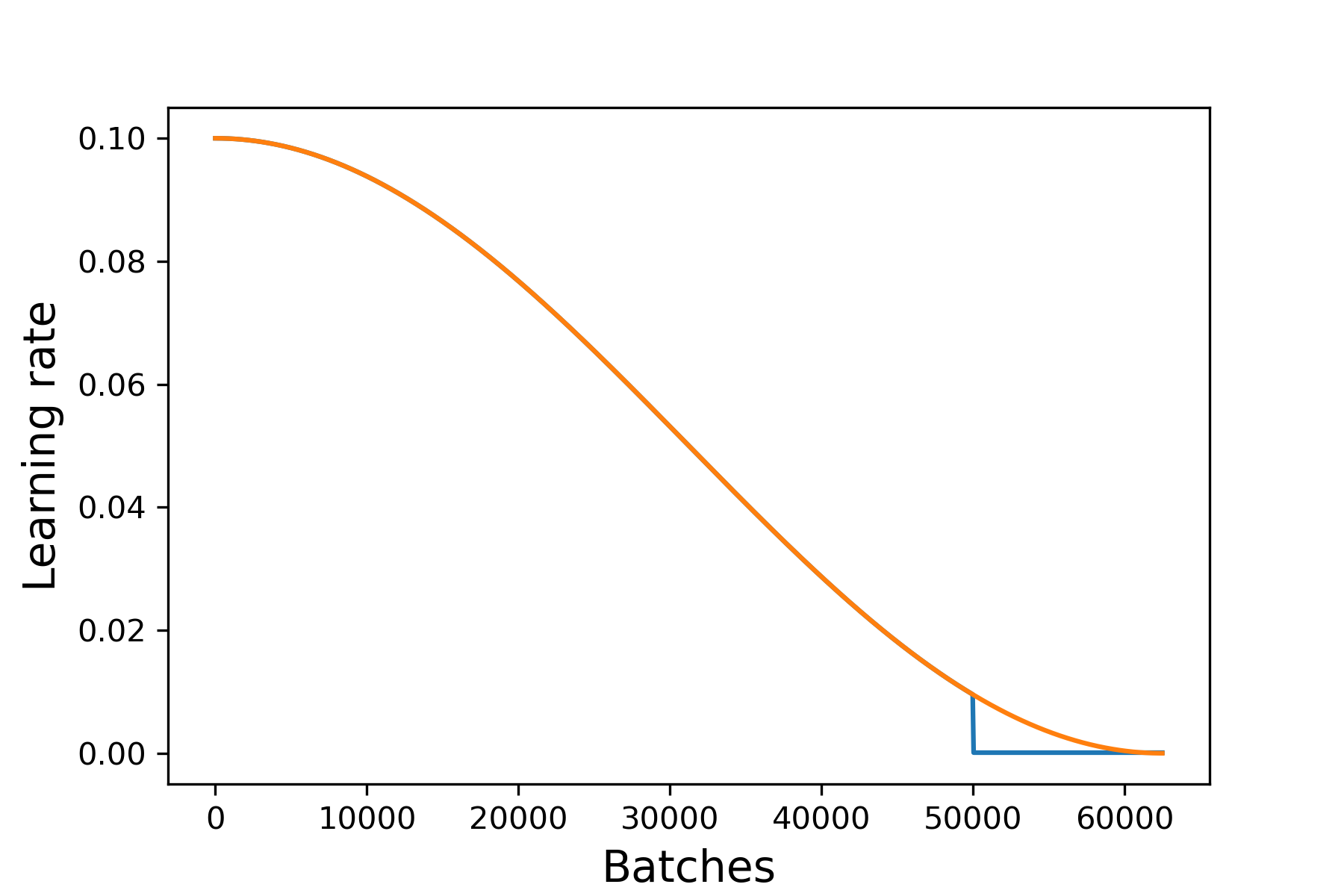}
\caption{Learning rate schedules.}
\label{fig:lr}
\end{figure}

We use stochastic gradient descent with no momentum as our optimizer.  We use 80\% of the dataset for training and reserve the remainder for validation to control early stopping.  We train all models (except for SGD trajectory) for 50 epochs using early stopping, with patience set to $10$ epochs.  The learning rate decreases for each batch.  We use a batch size of 32, so there are \numprint{1,250} batches per epoch.  For SGD trajectory, we run the decreasing learning rate schedule for at most 40 epochs (\numprint{50000} batches), then run $K=10$ epochs (\numprint{12500} batches) with a constant learning rate of $0.0001$.  For SGD trajectory, if we run into the early stopping criterion, we immediately jump to the constant learning rate stage.  The learning rate schedules are shown in Figure \ref{fig:lr}.  We set the initial learning rate to $r_0=0.1$.  We experimented with larger initial learning rates, but settled on one that worked well for all architectures\footnote{VGG13 did not work well with $r_0=0.5$, for example.}. 

We use existing implementations of ResNet20\footnote{\text{https://github.com/gahaalt/resnets-in-tensorflow2}}, VGG13\footnotemark[\value{footnote}], and DenseNet40\footnote{\text{https://github.com/titu1994/DenseNet}}.

\begin{figure*}
\centering
\includegraphics[width=6.5in]{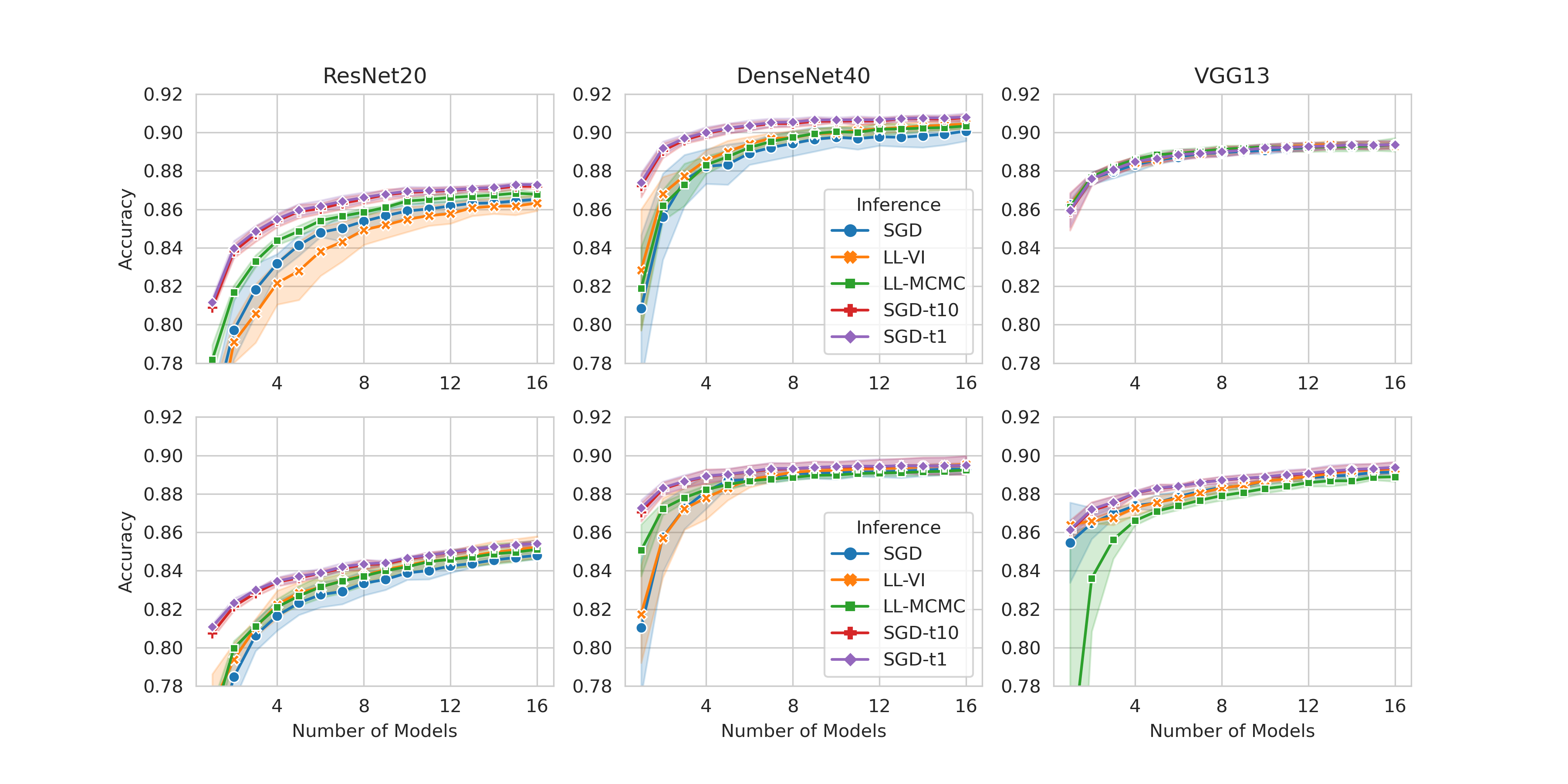}
\caption{Accuracy of models using different approximations to the BMA. The top row shows deep ensembles; the bottom row shows cyclic ensembles. The columns show three different network architectures. Each point represents the mean of five runs of the approximation method. The colored regions cover one standard deviation around the mean.}
\label{fig:accuracy}
\end{figure*}

\begin{figure*}
\centering
\includegraphics[width=6.5in]{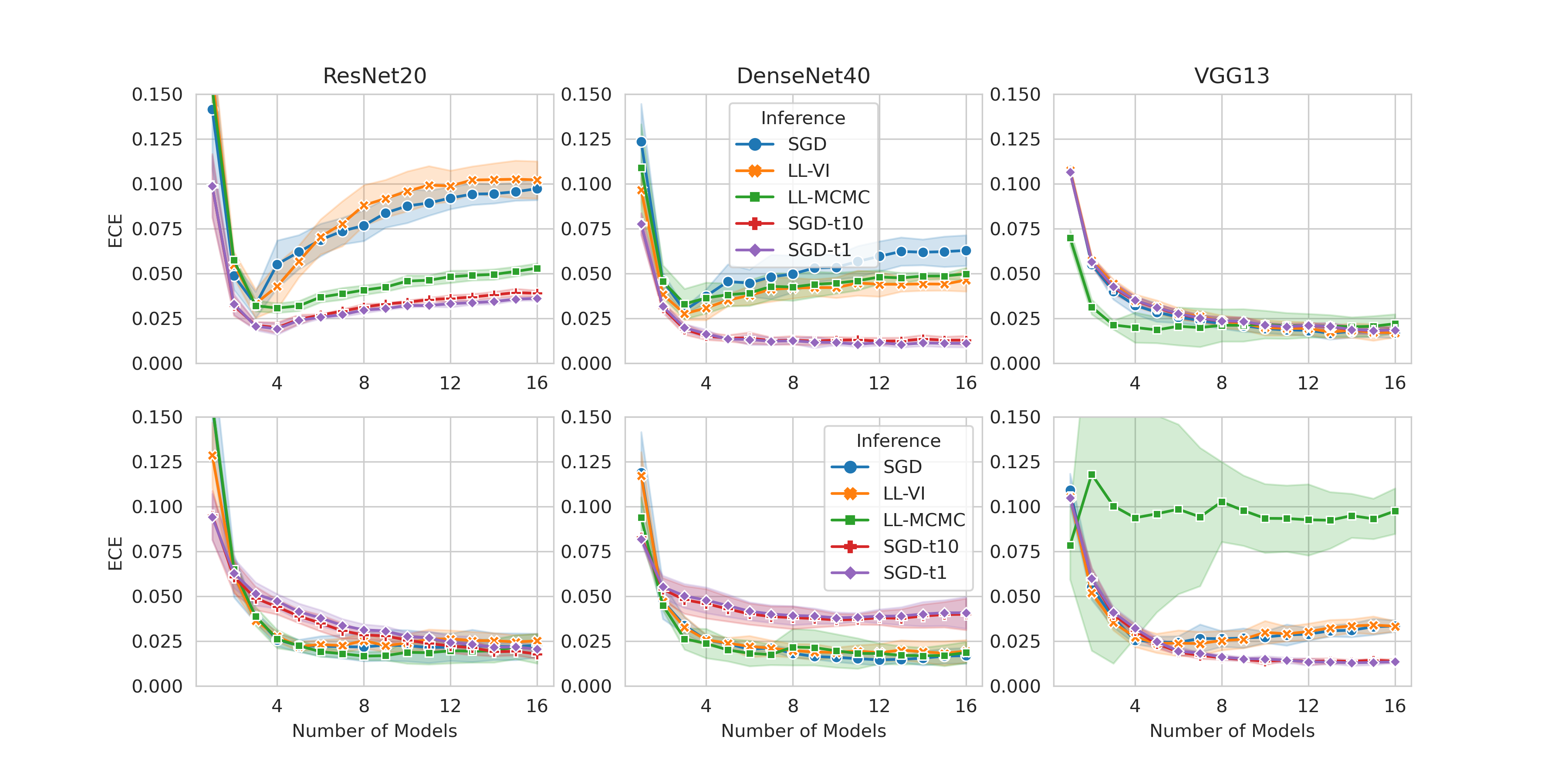}
\caption{Expected calibration errors of models using different approximations to the BMA. The top row shows deep ensembles; the bottom row shows cyclic ensembles. The columns show three different network architectures. Each point represents the mean of five runs of the approximation method. The colored regions cover one standard deviation around the mean.}
\label{fig:ECE}
\end{figure*}

\section{Acknowledgements}
DISTRIBUTION STATEMENT A. Approved for public release. Distribution is unlimited. This material is based upon work supported under Air Force Contract No. FA8702-15-D-0001. Any opinions, findings, conclusions or recommendations expressed in this material are those of the author(s) and do not necessarily reflect the views of the U.S. Air Force. © 2021 Massachusetts Institute of Technology. Delivered to the U.S. Government with Unlimited Rights, as defined in DFARS Part 252.227-7013 or 7014 (Feb 2014). Notwithstanding any copyright notice, U.S. Government rights in this work are defined by DFARS 252.227-7013 or DFARS 252.227-7014 as detailed above. Use of this work other than as specifically authorized by the U.S. Government may violate any copyrights that exist in this work.

\end{document}